# "I'm" Lost in Translation: Pronoun Missteps in Crowdsourced Data Sets

**KATIE SEABORN**

*Tokyo Institute of Technology*
*Tokyo, Japan*

**YEONGDAE KIM**

*Tokyo Institute of Technology*
*Tokyo, Japan*

**ABSTRACT:** As virtual assistants continue to be taken up globally, there is an ever-greater need for these speech-based systems to communicate naturally in a variety of languages. Crowdsourcing initiatives have focused on multilingual translation of big, open data sets for use in natural language processing (NLP). Yet, language translation is often not one-to-one, and biases can trickle in. In this late-breaking work, we focus on the case of pronouns translated between English and Japanese in the crowdsourced Tatoeba database. We found that masculine pronoun biases were present overall, even though plurality in language was accounted for in other ways. Importantly, we detected biases in the translation process that reflect nuanced reactions to the presence of feminine, neutral, and/or non-binary pronouns. We raise the issue of translation bias for pronouns and offer a practical solution to embed plurality in NLP data sets.

**KEYWORDS:** Gender bias, Data sets, Natural language processing, Translation, Feminist HCI

## 1  Introduction

Initiatives in natural language processing (NLP) for communication between people and virtual assistants (VAs) is taking off [9,28,36]. Many are gathering and offering large, crowdsourced data sets for this purpose [11]. An ongoing challenge is that most are in English and geared towards the US context [12,20,23]. In recognition of this, new efforts have started to be undertaken to ensure that a diversity of languages and cultural contexts are represented. For example, the Amazon Alexa team have been spearheading a "massive" crowdsourced translation and localization initiative of their MASSIVE data set into 51 languages [19], including a global competition[1]. Indeed, most of these initiatives rely on crowdsourcing activities. However, critical analyses have identified biases in translation, notably around gender representation in pronouns [32,41] and word choice [10,32,38]. As such, alongside the move towards voice- and speech-based interaction [15,34], gender bias in NLP is now a hot topic for feminist HCI work [3,4,6,13,32].

One issue that remains underexplored is how pronouns are *translated* language-to-language [14,21,30]. Languages vary greatly on pronoun use. For example, in English, a small set of pronouns are commonly used, but in Japanese, pronouns are less commonly used and arguably more varied [29]. We also need to consider gender ambiguous or "gendered ambiguous" pronouns that could represent more than one (or no) gender identity [41]. A common example is "they/them," which can be interpreted as a neutral plural pronoun or a singular non-binary pronoun. Moreover, trans*, non-binary, and other genderqueer folk may use "gender neutral" pronouns like "they/them" or gendered pronouns like "he/him" and "she/him," as well as a diversity of other pronouns, such as "ze/xe/xem/zem" [22,25,33,35]. Translators need to make careful choices about how to translate pronouns in a case-by-case fashion, and also be aware of the cultural context of variable pronouns. However, crowdsourcing contributors are likely not trained in such matters, as evidenced by the small body of work on pronoun translation for NLP data sets so far [32] and when considering related areas, such as crowdsourcing efforts and user-generated content for machine translation [14,21,30,42]. To the best of our knowledge, no work to date has looked at the context of NLP data sets geared towards VA training.

---

As a first step, we evaluated pronoun translations in *Tatoeba*[2], an open NLP database featuring crowdsourced translations between English and Japanese. We chose the case study of pronouns as an important identity, diversity, and inclusivity marker [25,31,33], as well as a pluralistic facet of language with implications for translation. Our overall research question (RQ) was: *Are there gender biases in pronouns within crowdsourced translations of English data sets to Japanese?* Our aim was to evaluate translation accuracy, so we first asked: *RQ1: Have pronouns been translated faithfully from English to Japanese?* We then focused on two important aspects when translating pronouns: (i) the relative number of pronouns linked to certain genders and (ii) whether one-to-one translations of pronouns were faithfully preserved. We were guided by long-standing language biases in English and romantic languages [5], recently confirmed within the latest crowdsourced English NLP data sets [8,17,18,32] and may thus feature in translations of these data sets to Japanese and other languages. As such, we asked specific questions about pronoun translation: *RQ2: Is there a masculine bias in pronouns across the data sets?* And *RQ3: Are there more instances of mistranslated masculine pronouns than other pronouns?* We aim to raise awareness of sociocultural biases that may be embedded in translation work and offer a technical solution to account for pronoun diversity and context-sensitive variability. We contribute our findings on the most recent and well-known NLP data sets used for training VA speech and a way to embed plurality in pronouns.

## 2   Related Work

We situate our work at the intersection of translations of NLP data sets and social justice initiatives within HCI.

### 2.1   GENDER MISTRANSLATIONS IN NLP

While virtually no work has considered gender translation biases in NLP data sets, a growing body of work has considered machine translation [14,21,30]. Google Translate, for instance, is a long-standing example of a gender biased system that relies on user-generated content; it specifically mistranslates pronouns and assigns masculine pronouns in the presence of ambiguous pronouns or a lack of pronouns [30,42]. Moreover, recent examples of mass-scale translation efforts in NLP spaces, specifically for VAs, indicates that biases, including gender biases, have not been

---

[2] https://tatoeba.org

considered. For example, the authors of the MASSIVE data set do not mention bias in their report[3]. As such, we hypothesize:

*H1. There will be mismatches in direct pronoun translations from English to Japanese.*

## 2.2   MASCULINE BIASES IN NLP

A wealth of research on NLP and machine translation has pointed to gender biases and especially masculine biases [10,12,20,32]. A recent word embeddings study of over 630 billion sources on the Internet evaluating the word "person" and "people" [2] found that these "gender neutral" words are often equated with "man" and "men." Breaking work by Seaborn, Chandra, and Fabre [32] has revealed implicit masculine biases in MASSIVE as well as the older ReDial NLP data sets, designed for training VAs and conversational user interfaces. This is not unprecedented, nor localized to NLP; this is a human language and cultural phenomenon. For instance, when there are no gender markers, such as names or titles, or pronouns, people are given to assume the person is a man [1]. We thus hypothesize for Tatoeba:

*H2. There will be a masculine bias in pronoun use in the data sets.*

Moreover, given the literature on crowdsourcing and user-generated content for machine translation alongside likely gender biases, we would expect contributors who make translations to carry over biases from one language to another. Indeed, a growing number of critical voices within NLP spaces have raised awareness of gender and other biases that can creep into the creation process, from who created the data to who curated and annotated it [7,27]. We expect this to be true for translation, which outside of NLP is recognized to be a process where biases come into play [14,39], including as a creative part of localizing and situating work within other cultures, i.e., transcreation [16]. As such, we hypothesize:

*H3. There will be a masculine bias in pronoun translation from English to Japanese.*

## 3   Methods

We conducted an automated content analysis of pronoun use in English and Japanese within Tatoeba. For this, we used NLP methods to extract meaningful patterns about pronoun presence and use, especially in translation, from the text materials in the data sets. Our protocol was registered before data collection on January 15[th], 2023 via OSF[4].

---

[3] https://arxiv.org/abs/2204.08582

## 3.1 DATA SET

We used the Tatoeba database of crowdsourced sentences and their translations across 359 languages, specifically the English and Japanese data sets. The downloadable version of the database is updated every Saturday at 6:30 a.m. (UTC); we used the version from December 24[th], 2022. While it is difficult to link translations to certain people, statistics indicate that contributors total 7,027 native[5] and 13,558 users[6] of English and 431 native and 2,997 users of Japanese. The data sets are made up of numbered rows, such that each row is paired with a row in the other language data set. Here is an example from the Japanese and English data sets:

> *Line 51: 僕が最後に自分の考えを伝えた人は、僕を気違いだと思ったようだ。*

> *Line 51: The last person I told my idea to thought I was nuts.*

Tatoeba was originally created by Trang Ho in 2006 and hosted on SourceForge as "multilangdict." Coincidentally, "tatoeba" is a Japanese word that means "for example." We chose Tatoeba for a few reasons. It is a top-ranked and downloaded resource at Hugging Face, a provider of NLP resources[7]. A diversity of sentences are continually being added to the database, which can be used to train and update the training of VAs and other NLP systems. As an open data set, it has a CC BY 2.0 FR licence, though some sentences are public domain. Notably, a portion of the English and Japanese sentences were sourced from the public domain Tanaka Corpus [40]. As a big data set, it has 255,675 sentences at the time of writing. It is an established and active crowdsourcing project, with participants around the world.

## 3.2 DATA PREPARATION AND ANALYSIS

All data analyses were performed in Python (version 3.8.10) with the Tatoebatools (version 0.2.1) package[8]. This package allowed us to download the Tatoeba data sets in Python. The English data set was tokenized, i.e., sentences were divided into separate words/terms, and tagged according to their linguistic meaning, e.g., nouns, pronouns, verbs, etc., using the Natural Language Toolkit (nltk) package (version 3.8.1). The Japanese data set was tokenized and tagged using the Janome[9] (version 0.4.2) package. For data preprocessing, we filtered out Japanese special characters, such as full-width spaces, '、 ，', ","," etc. Then we extracted pronouns from the English sentences using the pronoun tags created by nltk, namely the PRP and PRP$ tags, and the 代名詞 (dai-mei-shi) tag created by Janome from the Japanese sentences (refer to 3.3 for details). We identified the genderedness of the Japanese pronouns by referring to Jisho.org[10], which is an online Japanese dictionary that uses Tatoeba to provide example sentences between Japanese and English. We developed three pronoun categories: *masculine*, *feminine*, and, to account for pronoun ambiguity [41], *ambiguous*, including singular first-person pronouns, singular and/or plural "they/them" pronouns, words that may or may not be pronouns, such as 何 (nan; can refer to a person or "what"), and gender-diverse pronouns used by genderqueer people, such as "xim/xir," that could be pronouns or represent truncated words or typos [32], which only manual analysis could confirm.

---

[4] https://osf.io/8jmyc
[5] https://tatoeba.org/en/stats/native_speakers
[6] https://tatoeba.org/en/stats/users_languages
[7] https://huggingface.co/datasets/tatoeba
[8] https://pypi.org/project/tatoebatools

We generated descriptive statistics, i.e., means (M) and standard deviation (SD) of counts and ratios/percentages. We created a confusion matrix (or error matrix), which maps the relative distribution of one or more variables before and after algorithmic treatment [37] and is often used for automated content analysis and classification in machine learning (ML). We used inferential statistics, i.e., Chi-square tests, to compare groups, i.e., pronoun genders, languages, matching or not matching translations.

## 3.3   PRONOUN TAGS

We used the pronoun dictionary provided by the nltk package's Part of Speech (POS)[11] library for the English data set, and the tokenizer created by Kudo, Yamamoto, and Matsumoto [24] for the Japanese data set. However, because these libraries did not represent an up-to-date or inclusive spread of pronouns, we manually added several pronouns according to recent NLP work [25]. The lists of pronouns that we used are as follows:

**Japanese masculine:**

---

きみ、君、お前、俺、彼、彼ら、僕、君たち、オレ、おまえ、ぼく、ボク、僕ら、僕達、おれ、吾輩、キミ、てめぇ、小生、てめえ、僕たち

*kimi, kun, omae, ore, kare, karera, boku, kuntachi, ore, omae, boku, boku, bokura, bokutachi, ore, wagahai, kimi, temē, shōsei, temē, bokutachi*

**Japanese feminine:**

彼女、あたし、彼女ら

*kanojo, atashi, kanojora*

**Japanese ambiguous:**

何、私、それ、あなた、みんな、あいつ、誰、わたし、貴方、どなた、我々、あんた、そっ、やつ、みな、奴、あれ、なに、皆、みなさん、みなさま、我ら、余、彼等、だれ、奴ら、ウチ、わたくし、われわれ、よそ、われ、奴等、己、おのれ、何れ、わし、彼奴

*nani, watashi, sore, anata, minna, aitsu, dare, watashi, anata, donata, wareware, anta, so, yatsu, mina, yakko, are, nani, mina, minasan, minasama, warera, yo, karera, dare, yatsura, uchi, watakushi, wareware, yoso, ware, yatsura, onore, onore, izure, washi, aitsu*

**English masculine:**

*he, him, his, himself*

**English feminine:**

*her, she, herself*

**English ambiguous:**

*I, you, it, me, my, your, them, their, myself, they, themselves, we, us, oneself, our, yourself, its, itself, self, ourselves, 'em, theirs, thyself, one, ours, themself, theirself, theirselves, theirselves, xe, xem, xim, hir, xemself, xemselves, hirself, hirselves, ze, zeself, zeselves*

# 4   Results

We now turn to answering the hypotheses following data analysis.

## 4.1   H1. THERE WILL BE MISMATCHES IN DIRECT PRONOUN TRANSLATIONS FROM ENGLISH TO JAPANESE.

A Chi-square test was run to see if the number of pronouns by pronoun category (Table 1) differed between the English and Japanese data sets, i.e., because of the translation process, where pronouns were classified into three categories: masculine (EN = 43,453, JP = 50,350), feminine (EN = 19,025, JP = 17,123), and ambiguous (EN = 148,903, JP = 61,916). A medium, statistically significant relationship was found, $\chi2(2, N = 340,770) = 17,802.0$, $p < .001$, Cramer's $V = .23$. Overall, this confirms that a significant change occurred through the translation process, so we can accept H1.

**Table 1:** *Descriptive statistics by pronoun category and language.*

| Gender Category | English | Japanese |
| --- | --- | --- |
| Masculine | 43,453 | 50,350 |
| Feminine | 19,025 | 17,123 |
| Ambiguous | 148,903 | 61,916 |
| Non-Masculine | 167,928 | 79,039 |

## 4.2  H2: THERE WILL BE A MASCULINE BIAS IN PRONOUN USE IN THE DATA SETS.

Chi-square tests with Yates corrections were run to compare masculine and non-masculine pronouns by language. A statistically significant difference comparing masculine and feminine pronouns to ambiguous pronouns was found for the English data set, $\chi2(1, N=360,284) = 7930.40$, $p < .001$, Cramer's $V = .15$, and the Japanese data set, $\chi2(1, N=208,428) = 17,300.46$, $p < .001$, Cramer's $V = .29$. Descriptive statistics (Table 1) indicate that there were more masculine pronouns used regardless of language compared to feminine pronouns. Another Chi-square test was run to see if the number of masculine (EN = 43,453, JP = 50,350) and non-masculine (EN = 167,928, JP = 79,039) pronouns differed between the English and Japanese data sets. A weak, statistically significant relationship was found, $\chi2(1, N= 340,770) = 13,557.2$, $p < .001$, Cramer's $V = .20$. Descriptive statistics (Table 1) indicate that use of masculine pronouns increased in the Japanese data set, while use of other pronouns decreased. Overall, a significant difference in representation between masculine and non-masculine pronouns was found, apparently because of the translation process, supporting H2.

### 4.3   H3: THERE WILL BE A MASCULINE BIAS IN PRONOUN TRANSLATION FROM ENGLISH TO JAPANESE.

We considered whether the number of masculine (Match = 36,164, Mismatch = 21,475) and non-masculine (Match = 71,527, Mismatch = 103,913) pronouns differed based on translation mis/matching between English and Japanese. A Chi-square test revealed a low but statistically significant relationship between these two variables, $\chi2$(1, N = 233,079) = 8,426.7, $p$ < .001, Cramer's $V$ = .19. As a comparison, we then considered whether the number of feminine (Match = 16,855, Mismatch = 2,438) and non-feminine (Match = 90,836, Mismatch = 122,950) pronouns based on mis/matching between Japanese and English. mA Chi-square test revealed a low but statistically significant relationship, $\chi2$(1, N = 233,079) = 14,336.3, $p$ < .001, Cramer's $V$ = .25. The degree of feminine pronouns compared to other pronouns, in light of the results for masculine pronouns, indicates a higher interchange between ambiguous and masculine pronouns, i.e., a masculine bias, supporting H3.  Descriptive statistics (Table 2) indicate that use masculine pronouns decreased, i.e., were changed to other pronouns or removed on translation, thus leading us to reject H3.

**Table 2:** *Descriptive statistics for matching and mismatching translations.*

| Gender Category | Match | Mismatch |
| --- | --- | --- |
| Masculine | 36,164 | 21,475 |
| Non-masculine | 71,527 | 103,913 |
| Feminine | 16,855 | 2,438 |
| Non-feminine | 90,836 | 122,950 |

We created a confusion matrix (Figure 1) to clarify the interchange of English and Japanese pronouns, i.e., the extent to which each pronoun or combination of pronouns per sentence matched on translation. The diagonal values (in green) represent the cases where English and Japanese matched (only 57% of all pronouns). The top-right quadrant (in orange) represents the Japanese data, and the bottom-left quadrant (in purple) represents the English data. The largest difference was for ambiguous pronouns (English ambiguous, Japanese none = 73,228). The next largest difference was between English ambiguous pronouns and (i) Japanese masculine pronouns (n = 11,200) and (ii) Japanese sentences that include both masculine and ambiguous pronouns (n = 2,324). Both indicate a masculine bias in the Japanese data set. Comparatively, conversions between ambiguous and (i) feminine and (ii) sentences with feminine and ambiguous pronouns was only n = 96 and n = 50, respectively. Feminine pronouns scored the highest, which means that feminine pronouns were specified and excluded, while neutral and masculine pronouns were removed or used interchangeably. Finally, the matrix visualizes the masculine bias in matched cases between Japanese and English (n = 21,798).

| Matched Categories | Japanese | | | | | | | |
|---|---|---|---|---|---|---|---|---|
| | None | A | F | F,A | M | M,A | F,M | F,M,A |
| None | 60204 | 5171 | 39 | 5 | 129 | 21 | 1 | 0 |
| A | 73328 | 42657 | 96 | 50 | 11220 | 2324 | 25 | 3 |
| F | 1242 | 48 | 9771 | 409 | 2 | 0 | 12 | 0 |
| F,A | 421 | 221 | 1903 | 2471 | 51 | 7 | 372 | 19 |
| M | 4504 | 476 | 15 | 1 | 21798 | 1041 | 16 | 0 |
| M,A | 1230 | 827 | 5 | 1 | 5440 | 6012 | 8 | 3 |
| F,M | 76 | 5 | 101 | 10 | 57 | 6 | 1561 | 51 |
| F,M,A | 19 | 7 | 7 | 5 | 6 | 2 | 100 | 63 |

(Row label on left side: English. Legend on right: Highest / Lowest)

M: Masculine    F: Feminine    A: Ambiguous

**Figure 1:** *Confusion matrix showing the distribution of mis/matched pronoun translations. Counts by English pronoun categories (purple) are contrasted with Japanese pronoun categories (orange); matched cases fall along the diagonal line (green). Lighter tones indicate lower counts, while darker tones indicate higher counts. None: Non-gendered pronouns, i.e., "it," names, and non-humans.*

## 5  Discussion

Our findings reveal clear gender biases in pronouns for Tatoeba, notably masculine biases in the English and Japanese data sets. Moreover, there were mistranslations, and not only with respect to ambiguous pronouns. Furthermore, these mistranslations appear to have been driven by the relative presence and absence of other pronouns. We discuss these nuances and propose a fix for embracing pronoun plurality in Tatoeba and other NLP data sets.

### 5.1  NUANCES IN MISTRANSLATIONS: SOURCES OF MASCULINE BIASES AND PRONOUN MISSTEPS

A significant number of mismatching pronoun translations alongside an overrepresentation of masculine pronouns was found in Tatoeba, especially in the Japanese data set. Tatoeba translation can go either way, so we cannot be sure in which language these biases originated; we can only acknowledge the ground truth of the crowdsourced translation method itself. Even so, the results point to an increase in masculine pronouns in the Japanese data set, highlighted against the relatively low levels of feminine pronouns and mis/matched translations. One possibility is the presence of masculine language biases in Japanese [26]; yet, this is also true for English [2]. Without research directly comparing Japanese and English outside of translation and this small context of Tatoeba, it is hard to draw conclusions about whether we should expect either language to be more masculine-biased than the other. However, the case of ambiguous pronouns provides some nuance. Most were simply absent in the Japanese data set, likely due to the number of neutral pronouns such as "I" and "you" in English, which are omitted in Japanese, and "they/them," which can be a non-binary singular pronoun or a gender-neutral plural pronoun. Even so, the high number of masculine pronouns alongside ambiguous pronouns in the Japanese data set, where ambiguous pronouns are used in English, points to translation bias. Nevertheless, we can feel some confidence in the sheer number of ambiguous pronouns, which leave open the "genderedness" of language to the reader/listener and especially allow for queer readings of the text.

## 5.2    A PLURALISTIC APPROACH TO PRONOUN REPRESENTATION IN TRANSLATED DATA SETS

One of the most puzzling aspects of this work, particular to databases structured like Tatoeba, is the support of plurality for other language matters. Specifically, Tatoeba will repeat sentences in one or the other language to allow for a plurality of linguistic expressions, e.g., terms and grammar, and multiple translations of that sentence. For example:

> *Line 49: 生物学は好きになれません。*

> *Line 50: 私は生物学は決して好きではありませんでした。*

> *Line 49: I never liked biology.*

> *Line 50: I never liked biology.*

We might think to do the same for pronouns. However, there are a lot of pronouns, so this seems infeasible; the more lines added to the database, the more baseline repetition and the greater the file size of the database. We suggest another approach, drawing from similar efforts in NLP data sets: standardized labels or tags. In MASSIVE, for instance, there are tags for "time" and "date," as in the example below:

*{"id": "0", "locale": "en-US", "partition": "test", "scenario": "alarm", "intent": "alarm_set", "utt": "wake me up at five am this week", "annot_utt": "wake me up at [time : five am] [date : this week]", "worker_id": "1"}*

Here, the tags are merely annotating the data, rather than acting as variables that could replace the data. We suggest going one step further. A pronoun tag or placeholder would take the place of the text. More like a variable, it could refer to a list of pronouns appropriate for the sentence. To rework an example from Tatoeba:

*Line 6936:* 興味があるなんてものではなく、もう夢中なんです。

*Line 6936: [p] is not just interested, [p]'s crazy about it.*

There may be multiple lists depending on the context. For instance, there may be cases where a famous person or character with known pronouns is being referenced, and so to avoid misgendering that person, a special placeholder referring to certain subset of the pronoun list would be used. Future work can map out a formal specification.

We recognize that this would mean moving away from a plain text approach to a structured format, which we admit has implications for a project as large as Tatoeba; yet, we argue that, as a matter of inclusion and accuracy, this should be done. Still, we should consider how contributors would make contributions. Practically, it would be a simple matter of requiring a placeholder for pronouns rather than a particular pronoun within each sentence. Contributors could use a generic placeholder, such as [p] (for pronoun), which admins, corpus maintainers, and advanced contributors could then assess and revise, as needed. Still, we cannot expect to inform or train every contributor in a crowdsourcing initiative about new structures and rules. A future project could be the development of an AI-driven interface for inputting sentences into Tatoeba, one that is trained on the pronoun lists and watches for these pronouns, and either suggests to the contributor that they be replaced with a placeholder like [p] or automatically replaces them.

## 5.3  LIMITATIONS

We used one database (Tatoeba) and two languages (English and Japanese). While we provided reasons for this, we urge future work to evaluate whether and how the findings "translate" to other databases and other languages. Moreover, Tatoeba differs from other data sets in that the translation can go either way, so we could not determine the source of mistranslations. Future work will need to evaluate data sets with a known source language.

## 6    Conclusion

Human language is biased, and so is the human mind. We have revealed how such biases crop up in translations related to gender through the case study of pronouns in English and Japanese sentences from Tatoeba, an established, large-scale, open, crowdsourced translation project for NLP and used for training VAs. We aim to raise awareness, but we have also provided a fix that could be adopted by Tatoeba and other initiatives. Future work will need to explore the presence and extent of these "translation missteps" in other data sets and the feasibility of our proposed solution.

### ACKNOWLEDGMENTS

This work was funded in part by a Tokyo Tech Young Investigator Engineering Award.